\documentclass[10pt,journal,compsoc]{article}

\usepackage{hyperref}
\usepackage{color}
\usepackage{algorithmic}
\usepackage{algorithm2e}
\usepackage{graphicx} 
\usepackage{caption}
\usepackage{subcaption} 
\usepackage{geometry}

\begin{document}

\title{Guidelines for enhancing data locality in selected machine learning algorithms}

\author{Imen Chakroun, Tom Vander Aa and Tom Ashby\\
Exascience Life Lab, IMEC, Leuven Belgium\\
\{imen.chakroun, tom.vanderaa, tom.ashby\}@imec.be\\
}

\maketitle

\begin{abstract}
To deal with the complexity of the new bigger and more complex generation of data, machine learning (ML) techniques are probably the first and foremost used. For ML algorithms to produce results in a reasonable amount of time, they need to be implemented efficiently. In this paper, we analyze one of the means to increase the performances of machine learning algorithms which is exploiting data locality. Data locality and access patterns are often at the heart of performance issues in computing systems due to the use of certain hardware techniques to improve performance. Altering the access patterns to increase locality can dramatically increase performance of a given algorithm. Besides, repeated data access can be seen as redundancy in data movement. Similarly, there can also be redundancy in the repetition of calculations. This work also identifies some of the opportunities for avoiding these redundancies by directly reusing computation results. We start by motivating why and how a more efficient implementation can be achieved by exploiting reuse in the memory hierarchy of modern instruction set processors. Next we document the possibilities of such reuse in some selected machine learning algorithms. 
\end{abstract}



\paragraph{Notice}
This an extended version of the paper titled "Reviewing Data Access Patterns
and Computational Redundancy for Machine Learning Algorithms" that appeared in
the proceedings of the “IADIS International Conference Big Data Analytics, Data
Mining and Computational Intelligence 2019 (part of MCCSIS 2019)"
\cite{bigdaci19}

The final publication of this article is available at IOS Press through http://dx.doi.org/10.3233/IDA-184287.

\section{Motivation}

Because processor speed is increasing at a much faster rate than memory speed, computer architects have turned increasingly to the use of memory hierarchies with one or more levels of cache memory. This caching technique takes advantage of data locality in programs which is the property that references to the same memory location (temporal locality) or adjacent locations (spatial locality) reused within a short period of time. 

Increasing memory locality generally improves performance of a program significantly. One of the most popular ways to increase it is to rewrite the data intensive parts of the program, almost always the loops~\cite{looptrafos}. A simple example of this is to interchange the two loops in Algorithm~\ref{algo:before_interchange} such that the code looks like Algorithm~\ref{algo:after_interchange}; note that the indices in the loop headers have changed. If the matrices A and B are stored in column-major order, both the spatial and temporal reuse will be improved by the interchange because the same elements of B will be accessed in consecutive iterations.

\begin{algorithm}
  \SetAlgoLined
  \For{i = 1..N}
  {
    \For{j = 1..M}
    {
        A[i,j] = B[i-1,j] + B[i,j] + B[i+1,j]
    }
  }
  \caption{Before loop interchange} 
  \label{algo:before_interchange}
\end{algorithm} 

\begin{algorithm}
  \SetAlgoLined
  \For{j = 1..M}
  {
    \For{i = 1..N}
    {
        A[i,j] = B[i-1,j] + B[i,j] + B[i+1,j]
    }
  }
  \caption{After loop interchange} 
  \label{algo:after_interchange}
\end{algorithm}

Many such loop transformations exists and a large body of research has been dedicated to how and when to apply what transformation (see \cite{looptrafos} for an entry point to the literature). First and foremost the validity of the transformation is important, which is dictated by the dependencies in the
algorithm. Secondly one has to evaluate the benefit of a possible transformation, i.e. how it improves data locality, compared to how it affects other performance properties of the code. Several criteria need to be considered:

\begin{enumerate}
    \item Related to the data access pattern:

    \begin{itemize}
        \item What are the sizes of the different data structures in the loop?
        \item Are they only read or also written to?
        \item How many times is an element of the data structure reused?
    \end{itemize}

\item Related to the transformation:
    \begin{itemize}
        \item How does it affect the reuse of the different data
            structures? 
        \item How does it change the size of the different data structures?
        \item How does is change the complexity of the code?
    \end{itemize}
\end{enumerate}

\hfill

Machine learning algorithms and their implementations typically are computationally intensive and also operate on large data sets. To \emph{learn}, an ML algorithm needs to look at the \textbf{training} points (which together make up the training set). In supervised learning, which we consider in this report, the training set consists of data points for which the outcome to be learned is known: the points are \textbf{labelled}. Based on the training points, the \textbf{algorithm} builds a \textbf{model} by setting \textbf{model parameters} (also known as \textbf{weights}) so that the resulting model represents the underlying structure of the data. The model is evaluated on the \textbf{test} points to evaluate its accuracy. An algorithm and/or model may have extra parameters that are set by the user rather than learned from the data. These extra parameters are called \textbf{hyperparameters}. A particular algorithm being run on a particular data set to make a model is called a \textbf{learner}. To distinguish between particular hyperparameter settings for a learner we refer to a \textbf{learner instance} per hyperparameter tuple.

The train and test steps are typically performed several times for different types of model and different model hyperparameters e.g. by dividing between training and test sets. Hence the final program will be a set of \emph{nested loops} that could look like Algorithm~\ref{algo:loop_nests}.

\begin{algorithm}
  \SetAlgoLined
  \For{all learner types (1)} {
      \For{all free parameters combinations for the model (2)} {
          \For{all cross-fold combinations (3)} {
              \For{all folds in current combination (4)} {
                  \For{all samples in current fold (5)} {
                      Update current model based on current sample
                  }
              }
          }
      }
  }
  \caption{Example loop nest structure in a machine learning implementation} 
  \hspace{0.1cm}
  \label{algo:loop_nests}
\end{algorithm} 
\hfill

The \textbf{reuse distance} of a data location is the number of surrounding loop iterations that occur in between accesses to it. During training, most algorithms have to read each data point in the training set at least once. Such a full traversal of the training set is called an \textbf{epoch}. Once a learner is trained, the resulting fixed model can be used in an \textbf{operational phase} to make predictions for previously unseen points.

In the next sections we will investigate the potential for performance improvements using the language of loop transformations for code that looks like parts of Algorithm \ref{algo:loop_nests} by looking
at the criteria mentioned above. Unless otherwise stated, we are dealing with the training algorithms for a given learner rather than its use to classify unlabelled data in an operational environment. This is because the reuse available in the operational phase is often a) very little or zero and b) rather simple to exploit. Note that training also usually involves treating a portion of the available training data as a test set, and classifying the objects in the test set as though in an operational phase. Because test set objects may be a part of other training sets in some cases the reuse here is considered.

\section{Related work}

To the best of our knowledge, no similar research work centered on analyzing data access patterns and identifying locality and redundancy for ML algorithms exist. Indeed, conventional contributions dealing with reducing data access overheads in ML suggest using distributed approaches \cite{SHARED},\cite{CYCLADES}. In this case the focus is on single machine settings. We think however that such a review is a first step towards improving the performance and the efficiency in HPC settings.

\section{General Reuse}

This section deals with data reuse that is \emph{not} specific for one class of ML algorithms, but rather can be exploited in general. This means exploiting reuse in the outer loops of Algorithm~\ref{algo:loop_nests}.

For all opportunities that are described here, we explain what data structures are accesses and what is the potential for reuse. In the following the focus is on the possible redundancy and data flow that is related to the algorithmic evaluation that drive multiple executions of learners, e.g. cross-validation, conformal prediction, etc.

\subsection{Sub-sampling techniques}
\label{sec:sampling}

Sub-sampling techniques are frequently used in machine learning to estimate some property of a given learner, e.g. accuracy. They can also be used as part of model selection, when choosing a model based on the calculated property. The techniques work by sub-sampling the training set and running the learning
algorithms on the multiple sub-samples.

\subsubsection{Cross Validation}
\label{sec:crossvalidation}

Cross validation is an ML technique used to evaluate models in a way that takes possible model over-fitting into account. It can be used when evaluating different models resulting from learners derived from entirely different algorithms, or learner instances derived during the tuning of the hyperparameters of a model (or associated learning algorithm). Here we take the tuning of hyperparameters as a motivating example.

Many variants of cross validation exist, however, the most commonly known class is the $k$-fold cross validation. In this method, the training dataset is first divided into $k$ folds. For a given hyperparameter setting, each of the $k$ folds is considered as a testing set for evaluating the quality of the model which has been trained on the remaining $k-1$ folds. The overall performance of the model is the average of its performance on all $k$ folds. As a model selection technique, the hyperparameter setting can be chosen based on the best performance in terms of e.g. the prediction accuracy thus calculated. The pseudo code of the cross-validation technique is presented in Algorithm \ref{algo:cross_validation}.
   
\begin{algorithm}
  \SetAlgoLined
   \KwData{$T$ a set of training samples}
   Randomly divide the training dataset $T$ into $k$ (roughly equal sized) folds $f_i$\\
   \For{all $k$ folds $f_i$ in $T$ (1)} {
     Pick a fold $f_i$ as a test set.  \\
     Train the learner $L_i$ with $T \setminus f_i$  \\
     Test the learned model $M_i$ on $f_i$  \\
   }
   Estimate the accuracy of a learner based on the average performance across all folds.
   
  \caption{Pseudo code of the cross validation algorithm.} 
    \hspace{0.1cm}
  \label{algo:cross_validation}
\end{algorithm} 

\begin{figure*}
    \centering
    \includegraphics[scale=0.42]{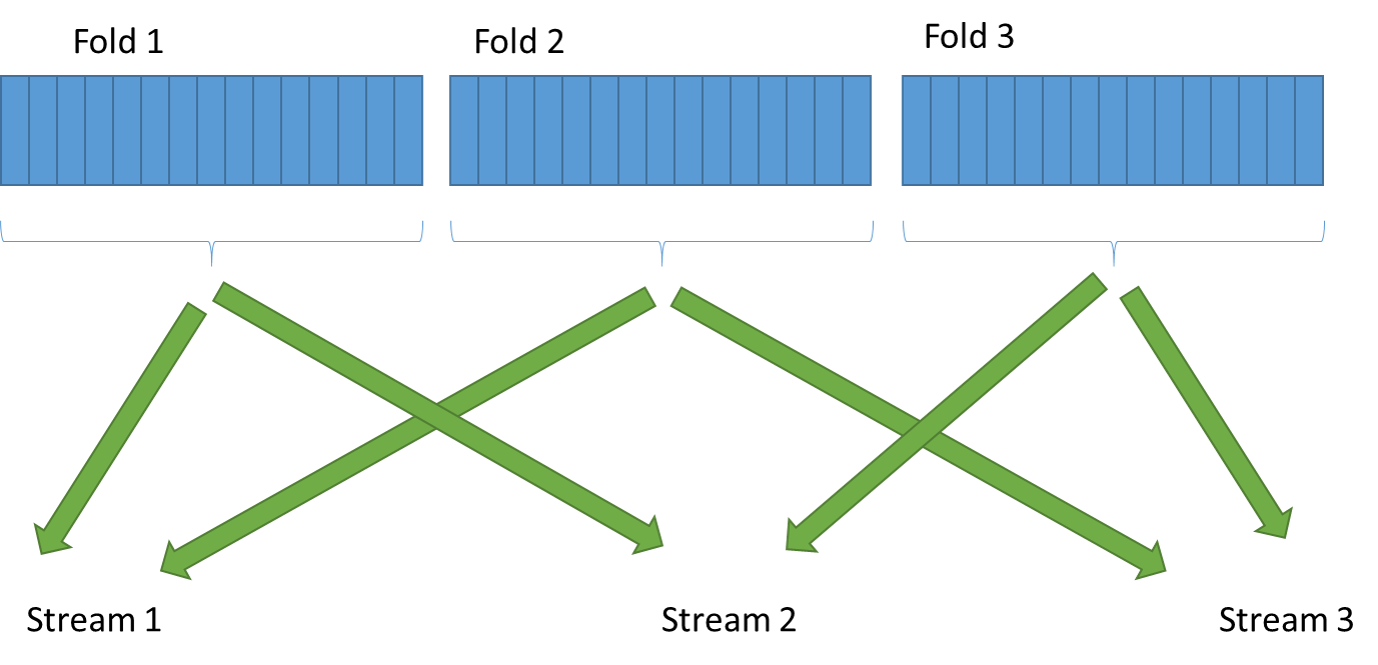}
    \caption{Folds feeding streams of points that are passed to instances of learners.}
    \label{fig:foldtostream}
\end{figure*}

The main external data read in this execution is the training set $T$. The set is read $k$ times, with the reuse being carried by the single loop at level 1. The reuse distance for each fold is 1 iteration of the outer loop. There is at least one data epoch per loop iteration (in the learning step). The simplest form of reuse for cross-validation is treating the different learner instances $L_i$ as black boxes and exploiting locality by passing the same fold to all the learners that need it simultaneously. This is illustrated in Figure \ref{fig:foldtostream}, where the different three folds contribute to three different streams of training points with each stream feeding one instance of a learner. This type of reuse can also be exploited for multiple completely different learners, all of which are undergoing cross-validation on the same training set $T$, which is equivalent to adding an outer loop over types of learner to algorithm \ref{algo:cross_validation}.

The model data for the learner instances is also iterated over, but that reuse is contained within the learning and testing steps.

\subsubsection{Bootstrapping}
\label{sec:bootstrapping}

Bootstrapping is similar to cross validation in the sense that is a statistical technique that involve re-sampling the data. However, the aims are quite different for each method. While cross-validation is often used to measure the generalisation accuracy of a learner, bootstrapping is more usually applied for evaluating the variance of the learned models\footnote{Cross-validation can be used to estimate variance but tends to be pessimistic, and bootstrap can be used to estimate accuracy but tends to be optimistic.}. Bootstrapping works as described in Algorithm \ref{algo:bootstrapping}.
 
\begin{algorithm}
  \SetAlgoLined
   \KwData{$T$ a set of training samples} 
   \For{$0$ to $n$ bootstrap samples} {
     Create a bootstrap  sample $B_i$ from the dataset $T$ by sampling with replacement \\
     Train the learner instance on $B_i$  \\
     Test the resulting model $M$ on the test set \\
   }
   Estimate the variance of the model based on the combined test results  \\
   
   \caption{Pseudo code of the bootstrapping algorithm.} 
     \hspace{0.1cm}
   \label{algo:bootstrapping}
\end{algorithm}
\hfill

The basic idea of the bootstrapping method is to take the original data set $T$ and generate a new equally sized sample from it called a bootstrap sample. The bootstrap sample is taken from the original using sampling with replacement (the same sample can be used several times within a bootstrap and between bootstraps). This is done a large number of times, meaning that each training point will appear in a significant number of bootstrap samples.

Similarly to cross-validation, the main data accesses that can be reasoned about are accesses to the training set, as sketched in algorithm template \ref{algo:bootstrapping}. Bootstrapping shows redundancy in multiple accesses to the same training point. Indeed, the inherent property of sampling with 
replacement in the bootstrapping algorithm means that a single sample can be encountered in different bootstrap samples and at different stages within the same bootstrap. Furthermore, the fact that a large number of bootstrap samples are often created means that the factor of reuse may be much higher than for
cross-validation. However, the access patterns are not quite so structured, meaning that arranging for the redundancy to be exploited may require significant effort or extra data structures. The number of full data epochs and the amount of reuse is less clear due to the statistical nature of the re-sampling. The reuse distance for most points will mostly be 1, and there may also be closer reuse than usual inside the training step due to re-sampling. 

Also, similarly to cross-validation, an outer loop can be added by considering multiple different types of learner. Access to the model occurs in the training and test steps. 

\subsection{Multiple Classifier Systems and Learner Selection}
\label{sec:ensembles}

Learner selection consists of comparing the performance of multiple learners and selecting the best for actual operational use in prediction. The algorithms can be completely different (e.g. $k$-NN and SVM) or be the same learner but differing only in some parameter setting as a result of parameter search. Evaluating different learners or learner instances this way on the
same training set immediately leads to reuse of that training set. If the training set can be accessed in the same order for the different learners, then this reuse becomes exploitable. This is essentially the same idea as applying loop interchange.

\begin{figure}
    \centering
    \includegraphics[scale=0.5]{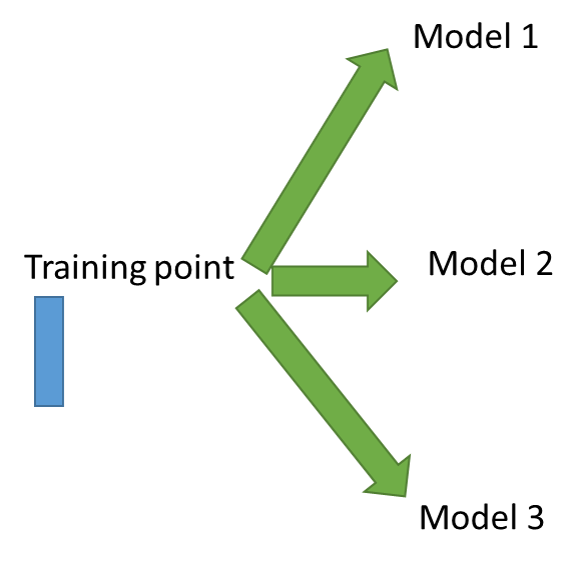}
    \caption{A point from a stream of training points being used for comparison with 3 different models from different learning algorithms.}
    \hspace{0.1cm}
    \label{fig:multmodel}
\end{figure}

This commonality of access order can happen across various levels of the algorithms. A coarse grain could be a the level of cross-validation folds or bootstrap samples. A finer grain would be visiting the same points driven by a common optimisation procedure (e.g. SGD); this idea is outlined in Figure \ref{fig:multmodel} where a training point is passed to be compared with multiple different models simultaneously. An even finer grain would be the level of the order of access to feature entries of a training point e.g. for gradient calculations (see Section \ref{sec:LR+SVM}).

Multiple classifier systems are built from multiple learners where the output of the learners is then combined in some way, e.g. by majority voting. This scenario is essentially the same as learner selection in terms of data access; each of the learners must still be individually trained. The data access is somewhat different for operational use though, in that classification inputs have to be passed through all the learners to get the final combined decision.

Ensemble learning takes this principle further in the case of having multiple learners produced by the same algorithm but with slightly different inputs \cite{ensemble_learning} or by multiple different learning algorithms. In this work, we consider two of the most popular ensembling methods, called \emph{bagging} and \emph{boosting}.

\subsubsection{Bagging} 

As its name indicates, bagging (bootstrap aggregating), is an ensemble method based on bootstrapping \cite{Breiman1996}. While bootstrapping is a sampling technique used for estimating statistical variance of a single learner, bagging is a machine learning ensemble method which considers several learners. The
bagging technique proceeds as follows: from an original training set, multiple different training sets (bootstrap samples) are generated with replacement. For each bootstrap sample, a model is built. All models are combined in an ensemble of models that are used for the test set and a majority vote is returned as a result. The bagging training algorithm is given in \ref{algo:bagging}.

As a method inherently derived from bootstrapping, the same data reuse opportunities identified in \ref{sec:bootstrapping} are valid for the bagging technique. 

\begin{algorithm}
  \SetAlgoLined
   \KwData{$T$, a set of training samples} 
   \For{all learner instances (1)} {
     Create a bootstrap  sample $B_i$ from the dataset $T$ by sampling with replacement. \\
     Train the learner instance on $B_i$.  \\
   }
   Keep the learner instances to vote on unlabelled points.
   \caption{Algorithm template of the bagging technique} 
   \hspace{0.1cm}
   \label{algo:bagging}
\end{algorithm}

\subsubsection{Boosting} 

The basic idea of boosting is similar to bagging since boosting also creates an ensemble of classifiers by re sampling the data, which are then combined by majority voting. The difference between both is in the re sampling step which is done without replacement in the boosting technique. 

 \begin{algorithm}
   \SetAlgoLined
    \KwData{$T$, a set of training samples} 
   	\For{all learner instances (1)} {
 			 Create a sample \emph{S$_1$} from the dataset \emph{T}; \\
 			 Train the learner instance M$_1$ on \emph{S$_1$}; \\
 			 Evaluate the accuracy of M$_1$ on the test set; \\
 			 Generate a sample \emph{S$_2$} randomly from \emph{T} such that for half of the samples M$_1$ provides correct predictions, and for another half incorrect ones; \\
 			 Train learner instance M$_2$ on \emph{S$_2$}; \\ 
 			 Train M$_3$ on samples from \emph{T} where M$_1$ and M$_2$ classify differently; \\			
 			 Test M$_1$, M$_2$ and M$_3$ on the test set and decide according to a majority vote; 
              }
   \caption{Algorithm template of the boosting technique.} 
   \hspace{0.1cm}
   \label{algo:boosting}
 \end{algorithm}%

The boosting algorithm template given in algorithm \ref{algo:boosting} shows that boosting proceeds by constructing the most informative training data for each consecutive classifier. Indeed, each iteration of boosting creates three classifiers: the first model M$_1$ is trained with a random subset of the available training data. The training data subset for the second model M$_2$ is chosen as the most informative subset, given M$_1$. Practically, M$_2$ is trained on a training data constructed such that only half of it is correctly classified by M$_1$. The third model M$_3$ is trained with instances on which M$_1$ and M$_2$ disagree. The three models are combined through a three-way majority vote. 

One opportunity for increasing data reuse is here to compute the cost function of some samples being part of two or three of the models M$_1$, M$_2$ and M$_3$ only once and use the results whenever needed.

\subsection{Optimisation} 
\label{sec:optimisation}

The training or fitting of many models is based on optimisation techniques, with many approaches based on gradient based optimisation, using either first or second order gradients. Due to the expense of gradient calculations for complex models on large data sets, an approximation is often used, hence we consider these algorithms here.

\subsubsection{Stochastic Gradient Descent} 
\label{sec:SGD}

Many machine learning algorithms such as logistic regression, support vector machine, neural networks, etc.. can be formulated as a mathematical optimization problem where the key challenge is to identify a function that best generalizes from the training data and to optimize the parameters of it so that the number of miss-classified points is minimized. The learned function is used in the operational phase
to make the predictions for the unlabelled points. In order to capture the quality of the model, a loss function calculating the error between the predicted and the real values on the training set (and some regularisation parameters) is used. The further away the prediction is from the real value (for some metric) of a training point, the larger the value of the loss function is. The goal of training the model is therefore to minimize the underlying loss function (also called cost function).  For achieving this, Stochastic Gradient Descent (SGD) is a quite popular algorithm. It is an optimization method that attempts to find the values of the model coefficients (the parameter or weight vector) that minimizes the loss function when they cannot be calculated analytically. SGD has proven to achieve state of-the-art performance on a variety of machine learning tasks \cite{bottou_2012,sgdBPMF}. With its small memory footprint, robustness against noise and fast learning rates, SGD is indeed a good candidate for training data-intensive models. 

SGD is a variety of the gradient descent (GD) method \cite{gradient_descent} with a major difference in the number of updates per visited data point. In each iteration, GD sweeps through the complete training set to calculate an update to the weights vector. For SGD, only a random element from the training data is considered every iteration to perform the update of the weights vector. A template of both methods is
described in Algorithm \ref{algo:GD} where the input parameter $n$ defines the number of points to consider in the gradient computation and hence the algorithm. For SGD, $n$ is equal to 1 while for GD the $n$ is equal to the size of the complete training set.

\begin{algorithm}
  \SetAlgoLined 
  \KwData{$T$, a training set\\
  $M$, an initial guess for the model parameters\\
  $n$, the batch size for calculating the gradient} 
      
    \For{The required number of epochs} {
      \For{Each batch $B$ of size $n$ in $T$} {
        \For{Each training point $t$ in $B$} {
    	  Compute the gradient $g$ of the loss function for $M$ with respect to $t$ \\
          Add $g$ to the combined gradient $G$
        }
        Update model parameters $M$ by a step in direction $G$ \\
        Update algorithm parameters such as step size etc.
      }
    }
    \caption{(Stochastic) Gradient descent algorithm template.} 
    \label{algo:GD} 
\end{algorithm} 

The main data touched in both algorithms is the training set, the model and the calculated gradient. The reuse distance for any training point in both algorithms is $|T|$, the size of the training set. Similarly, any calculated gradient $g$ is used directly in the same iteration (reuse distance 0), and the model is reused every iteration (reuse distance 1). The learning rate parameter is updated as frequently as the weight vector for each algorithm. 

One of the main disadvantages of GD compared to SGD is that it requires many epochs before reaching a reasonably good solution and thus requires more data movement which is expensive. It is also computationally heavier since the update is calculated based on the whole training set.  Even if
in common machine learning problems the objective function to minimize is only a cumulative sum of the loss over the training examples, with the increasing size of the data handled in big data use-cases the overheard remains expensive. SGD is a popular, cheaper alternative which provides $|T|$ model updates per epoch, rather than 1.
 
Due to its "stochastic" characteristic, using SGD can introduce a lot of noise in the gradient during its search for a minimum and the path towards the eventually discovered minimum is not as smooth as it is in GD. A common best practice is therefore to use a third algorithm called Mini-Batch Gradient
Descent (MB-GD) which aims at converging as quickly as SGD and as smoothly as GD.  In MB-GD, the model is updated based on small groups of training samples called mini-batches, as shown in Algorithm \ref{algo:MB-GD}. For this reason, MB-GD converges in fewer iterations than GD because we update the weights more frequently. For the same reason, using MB-GD reduces the noise introduced by computing the gradient using one random sample at a time because the accuracy of the direction and the step to take is higher when more points are used, an thus may also converge faster than SGD. The data touches and reuse distances are the same as for GD and SGD, but now there are $|T|/n$ model updates per epoch. Note that the mini-batch size is an algorithm hyperparameter.

Gradient descent-like optimization is often used with several learners. The data traversal and the number of data touches is largely determined by the optimization algorithm used \emph{regardless of the model being trained}. Hence, an intuitive idea for a data reuse-aware coding is to fold different models together and train them simultaneously using the same optimization method, thus re-using the stream of
data. 

\begin{algorithm}
  \SetAlgoLined
  \KwData{$T$ a set of training samples \\
  $M$ an initial guess for the model parameters} 

  Randomly shuffle the order of all the training data in $T$\\
  Divide $T$ into mini-batches of size $n$

  \For{The required number of epochs} {
    \For{Each mini-batch $b$ in the epoch} {
      \For{All training points $t$ in $b$} {
        Compute the gradient $g$ of the loss function for $M$ with respect to $t$ \\
        Add $g$ to the combined gradient $G$
      }
      Update the weights by a step in direction $G$ \\
      Update algorithm parameters such as learning rate etc.
    }
  }
  \caption{MB-GD algorithm template}
  \label{algo:MB-GD}
\end{algorithm} 

\section{Algorithm-specific Reuse}
\label{sec:specific}

This section identifies data reuse opportunities specific to first and foremost used and known classes of supervised ML algorithms. We particularly focus on Instance-based learners, Naive Bayes learners, Logistic regression and SVM and Neural networks.

\subsection{Instance-based learners} 
\label{sec:IBL}

Instance-based learning (also known as memory-based learning) is a class of ML algorithms that do not construct a model as in usual classification methods. They are called instance-based because they construct the hypotheses directly from the training points themselves, and the learner is ``memorising'' the training set. In several of these methods no training phase takes place (with the possible exception of a search for the best value of some hyperparameter and the predictions are made by measuring similarities between points to predict and points from the training
data set. The similarity measure and way of combining input from the training points defines the different types of instance-based learners. 

\subsubsection{K Nearest Neighbours}

The principle behind the $k$ nearest neighbour ($k$-NN) method is to find the $k$ training samples closest in distance to the new point, and predict the label from these. The distance can, in general, be any metric but the standard Euclidean distance is definitely the simplest and most used technique (if the attributes are all of the same scale). An unknown instance point is classified by a majority vote of its neighbours. For a classification problem, the output of $k$-NN is a class membership while for a
regression problem, the output of $k$-NN is the average of the values of its $k$ nearest neighbours. The algorithm template of the $k$-NN technique is given in Algorithm \ref{algo:KNN}, but note that this is the classification phase, not the training phase as this instance based learner is not trained.

\begin{algorithm}
  \SetAlgoLined
  \KwData{$RT$, the set of remembered training points\\
  $P$, the set of points to predict}
  \For{all $i$ points in $P$ (1)} {
    \For{all $j$ remembered training points in $RT$ (2)} { 
      $d = compute\_distance(i,j) $\\
      \If {$d \leq$ the $k$ closest distances} {
        Add $j$ to the list of $k$ nearest neighbours of $i$\\
        (Remove further away members from the list if necessary)
      }
    }
    \For{all $j$ in the list of $k$ nearest neighbours of $i$} {
      Add class of $j$ to the running vote count for $i$
    }
    Return the class of $i$ based on the majority vote
  }
  \caption{Algorithm template of the $k$ nearest neighbours method (classification).} 
  \label{algo:KNN}
\end{algorithm}

The data accessed in algorithm \ref{algo:KNN} are the points in $RT$ and the points in $P$. The point from $P$ being classified is reused directly in each iteration of loop level 2, with a reuse distance of one. The reuse of training points from $RT$ is carried by loop level 1, with reuse distance $|RT|$. Each execution of loop level 2 is an epoch. There is some reuse in the handling of the list of $k$ nearest
neighbours for any point, but this is likely to be a trivial amount of data and computation compared to handling $RT$ and computing distances.

The only simple way to improve reuse is to shorten the reuse distance for elements of $RT$ by calculating distances to multiple prediction points simultaneously; an appropriate batch size can be calculated based on cache sizes available. Note that the approach to $k$-NN presented here is a very simple one. There are various more sophisticated approaches available (e.g. \cite{FNNpackageinR}), which are related to approaches to kernel matrix approximation. 

Despite there being no training phase as such for many instance based learners such as $k$-NN (as the model has no parameters to set), hyperparameter optimisation, such as searching for a good value of $k$, can be be thought of as a form of training. In this case the accuracy of any given choice of hyperparameters will be evaluated using e.g. cross-validation, with the aim of then picking the
best. This then leads to patterns of reuse. Using $k$-NN inside a cross-validation procedure would add an outer loop where each test point from a given fold $f_i$ is classified based on the mutual distances to $RT \setminus f_i$. This leads to both redundancy of access and redundant computations, in
that the same mutual distances will be repeatedly calculated. 

\subsubsection{Parzen-Rosenblatt Window}

The second instance-based ML algorithm considered here is the Parzen-Rosenblatt window (PRW) density estimation method which has been introduced by Emanuel Parzen and Murray Rosenblatt respectively in \cite{Parzen} and \cite{Rosenblatt}. Its basic idea for predicting the probability density function $P(x)$ is to place a window function $f$ at $x$ and determine what are the contribution of each training point $x_i$ to the window. The approximation to the probability density function value $P(x)$ is computed using a kernel function that returns a weighted sum of the contributions from all the samples $x_i$ to this window.

PRW has two important hyperparameters, the window bandwidth and the kernel function. A great number of bandwidth selection techniques exist such as \cite{Jones1996, Sheather2004}. For the kernel function, different variants exist: Gaussian, Epanechnikov, Uniform, etc. The Gaussian probability density function is the most popular kernel for Parzen-window density estimation because it has no sharp limits as the kernels listed above, it considers all data-points and produces smooth results with smooth derivatives.

From an implementation point of view, as sketched in the Algorithm \ref{algo:PRW}, computing PRW implies evaluating a large number of kernel-functions at a large number of data points. Given that the kernel function similarities are often based on Euclidean distance, the similarity with $k$-NN is obvious. The data reuse and distances involved are hence the same as for $k$-NN. The same optimizations are also applicable for the $k$-NN method as mentioned above.  

\begin{algorithm}
  \SetAlgoLined
  \KwData{$RT$, the set of remembered training points\\
  $P$, the set of test points to predict}

  \For{all $i$ test points in $P$ (1)} {
    \For{all $j$ remembered training points in $RT$ (2)} { 
      $s = compute\_similarity(i,j) $\\
      Add $s$ to the running total for the $C$, the class of $j$\\
    }
    Return the class of $i$ based on the class with the highest total weight $C$\\
  }  
  \caption{Algorithm template of the Parzen-Rozenblatt Window method.} 
  \label{algo:PRW}
\end{algorithm}

\subsection{Naive Bayes learners}
\label{naivebayes}

Naive Bayes methods are a set of supervised learning algorithms based on applying Bayes' theorem with the assumption of independence between every pair of features (the occurrence of a particular feature in a class is unrelated to the presence of any other feature). Because it does not include iterative
parameter search, a Naive Bayes model is easy to build (not very CPU and memory consuming) and particularly useful for very large data sets. 

The naive Bayes model calculates a posterior probability $P(c|x)$ of a class $c$ given a feature $x$ using the equation \ref{eq:eq1} (Bayes Theorem) : 

\begin{equation} \label{eq:eq1}
P(c|x) = \frac{P(x|c) P(c)}{P(x)}
\end{equation}

Where:
\begin{itemize}
\item $P(c|x)$ is the posterior probability of class $c$ given the predictor x.
\item $P(c)$ is the prior probability of class $c$.
\item $P(x|c)$ is the probability of predictor $x$ given class $c$.
\item $P(x)$ is the prior probability of predictor $x$.
\end{itemize} 

In practice $P(x)$ is ignored as it is only a normalising constant. As illustrated in the algorithm template \ref{algo:BAYES}, the training phase of naive Bayes iterates over all the training points and over all the features of each training point in order to calculate the occurrence of each feature per
class and the occurrence of each class.

During the testing or operational step, the naive Bayes learner loops over all the points to predict,
over all the attributes of every test point and over all the available classes and calculates the (numerator of the) posterior probability of each given each test point using the pre-computed structures aforementioned. These posterior probabilities are sorted and the best score determines the class returned for every instance point.

\begin{algorithm}
  \SetAlgoLined
  \KwData{$T$, a training set}
  \For{all the features $x_{i}$ (1)} {
    \For{all classes $c$ in $C$ (2)} {
      \For{all training points $t$ in class $c$ (3)} {
        Calculate the distribution of attributes for that class, $P(x|c)$, by fitting the distribution model (e.g. a Gaussian) to the feature values
      }
      Keeping a running total of the number of training points in class $c$
    }
  }
  Calculate $P(c)$ based on the class sizes \\
  Return the most probable class
  
  \caption{Algorithm template of the Naive Bayes algorithms.} 
  \hfill
  \label{algo:BAYES}
\end{algorithm}

The main data touched are the training points $T$. There is some partial reuse of training points carried by loop level 1. For each feature, the information for that feature is read only once, so there is no reuse of any individual feature in any given training point. However, spatial hardware cache locality may result in there being some ``accidental'' quasi-reuse of parts of the training vectors. The quasi-reuse is carried by loop level 1 and has distance $|T|$. The model is trained with only one epoch.

As for other learners, reuse arises carried by loop level 1 if the naive Bayes is used with the sampling or ensembles techniques discussed in Section \ref{sec:sampling} and \ref{sec:ensembles}.

\subsection{Logistic Regression and Support Vector Machine} 
\label{sec:LR+SVM}

Logistic Regression (LR) and Support Vector Machine (SVM) are two closely related approaches for fitting linear models for binary classification. The optimum placement of the hyperplane dividing the two classes is determined by the loss function, which determines the per training point contribution to the
total cost of any given model parameters, or the gradient of the cost which can then be used for optimisation. Optimisation is made easier by the fact that the loss function is convex.

In terms of the access to data, the two approaches are essentially the same. For a given batch of training points, the distance of the point to the model hyperplane is calculated with an inner-product per training point to produce a scalar, which is then fed into the differentiated loss function (to
calculate the gradient). For the purpose of discussion we assume an Euclidean distance and the associated inner-product. Algorithm \ref{algo:linearModelGrad} describes the procedure (the bias term is omitted for clarity). 

The cost function for both LR and SVM (and any other type of model that is being fitted using optimisation) usually also has a regularisation term. Regularisation is typically phrased as some sort of additional cost on the weights in the model to stop them growing too large, or to attempt to send as
many weights as possible to zero (sparsification). The differentiated version of the standard regularisation term is simple enough that it can be phrased as \emph{weight decay}. If the gradient being calculated is not sparse, then the main data use and calculation in the update of the model is the calculation of the gradient and, to a less extent, the update of the model based on it. Otherwise applying weight decay at each step may be more expensive due to the complete traversal of the model.

Both types of model can be effectively trained using variants of gradient descent such as SGD (see \ref{sec:SGD}) as implied by the update calculation based on a gradient computed with respect to a mini-batch. For SVMs, this is known as training the primal form of the problem.

The reuse in a single batch update for these algorithms is relatively limited. Each training point $t$ in the batch is accessed only once. The majority of accesses to the model $M$ is carried by loop $1a$; there is one inner-product per training point. This leads to a reuse distance of $|M|$. These loops can be easily rearranged to carry out the inner-products simultaneously so that there is one traversal over the model for those operations. The update of the weights in loop $1b$ is, in the basic algorithm, dependent on the gradient calculated in loop $1a$, so these two calculations are effectively sequentialised.

\begin{algorithm}
  \SetAlgoLined
  \KwData{A batch of training samples $B$ \\
    An initial guess for the hyperplane model $M$ \\
  } 
  
  \For{all training points $t$ in $B$ (1a)} {
      \For{all vector indices $i$ in $t$ and $M$ (2)} {
        Multiply scalars $t_i$ and $M_i$ \\
        Accumulate in $p$ (the inner product)\\
      }
      Accumulate the gradient $f'(p)$ in $g$ (the batch gradient)\\
  }
  \For{all weights in model $M$ (1b)} {
    update the weight based on weight decay, training schedule step size and the entry in $g$\\
  }
  Return $M$\\
  \caption{Pseudo-code for updating a linear model with a mini-batch based (gradient descent)} 
  \hfill
  \label{algo:linearModelGrad}
\end{algorithm}

If these two algorithms are to be run on the same training set note that they can be quite tightly coupled. The training points can be presented to both models in the same order, meaning that the common visit to a training point can be combined so that the data only needs to be touched once. Indeed, the
inner-product of the training point with the different hyperplane models can be done at the same time so that there is direct reuse in a feature-by-feature way of the training point.

\subsection{Neural networks} \label{sec:ANN}

Artificial neural networks (ANNs) are graph models inspired by biological neural networks. Each node in the graph is an artificial neuron, which consists of some inputs with associated weights, a scalar non-linear activation function, and a number of outputs. For the feed-forward neural networks that we consider here, neurons are arranged as a Directed Acyclic Graph (DAG) with groups of neurons in layers connected to the layer either side. 

Weight gradient calculation for ANNs is done with the back-prop (backwards propagation) algorithm. Finite differencing is not used due to the large number of weights to optimise. Back-prop consists of three phases, a forward propagation pass, a backward propagation pass and a gradient calculation. Forward propagation consists of presenting an input to the network and calculating the network output for that input whilst saving various intermediate values, namely the output of all neurons in the network and the weighted input to all activation functions. Backward propagation consists of propagating the error backwards through the network, using operations that are similar to forward propagation. The error for a given neuron is the rate of change of the cost with respect to the rate of change of the weighted input for this neuron. This value is used with the neuron inputs (saved from forward propagation) to calculate the actual gradient of individual weights in gradient calculation, the final step. 

Training for such models is usually done with SGD (or a variant), based on mini-batches of training samples, see Section~\ref{sec:optimisation}. The main cost of the optimisation for an ANN is the calculation of the gradient at each step. Such a mini-batch can be represented as a matrix with columns corresponding to network input vectors.

\subsubsection{Forward and Backward Propagation Sweep} 

\paragraph{Forward Propagation:} In more detail, the forward propagation step is shown in algorithm \ref{algo:forward-prop}. The external data read in the forward pass is the network weights (i.e. the model) and the input to the first layer. This output of each layer $a$ is produced and stored, and also used as input for the following layer or as the final network output. The weighted sums $z$ are also produced and stored.

\begin{algorithm}
  \SetAlgoLined   
  \KwData{$B$, a mini-batch of training points\\
  $M$, a model}
  \For{all layers in order (1)} {
    \For{each set of intermediate values from the mini-batch $B$ (2)} {
      \For{all neurons in the layer (3)} {
        \For{all weights $w_i$ for this neuron (4)} {
          Multiply the input from the previous layer $a_i$ by the weight $w_i$\\
          Add the result to the running total of the weighted inputs $z$\\
        }
        Record the total weighted input $z$ for later use\\
        Apply the activation function to the total weighted input to get the output $a$\\
        Record the neuron output $a$ for later use\\
      }
    }
  }
  \caption{Algorithm template for neural network forward propagation} 
  \label{algo:forward-prop}
\end{algorithm}

\begin{figure}
     \centering
 \includegraphics[scale=0.5]{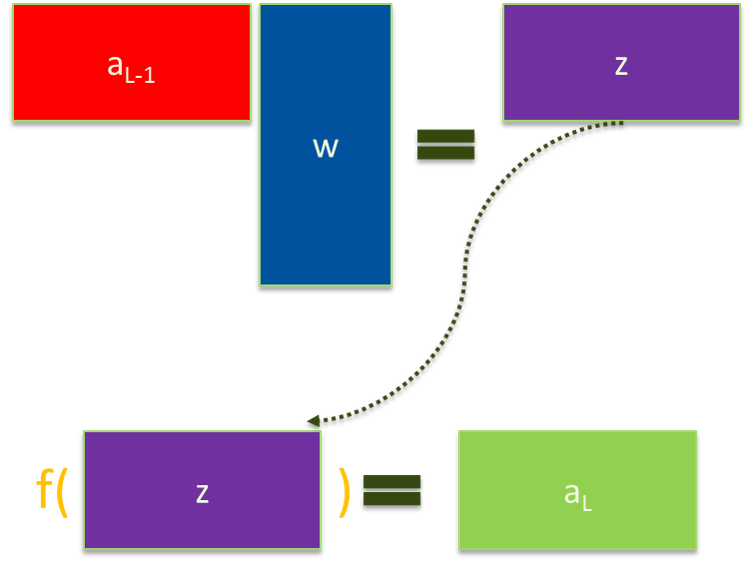}
 \caption{Calculating the feed forward of the mini-batch output of NN layer $L$ as matrix operations. The collection of output vectors from layer $L-1$ for the mini-batch is in matrix $a_{L-1}$. This is multiplied by the matrix of weights $w$ for all neurons in layer $L$ to give $z$, the collection of weighted input vectors for $L$. Applying activation function $f()$ point-wise to $z$ gives the collection of mini-batch outputs for this layer, $a_L$.}
 \label{fig:feedforward} 
\end{figure}

In the forward propagation pass, the direct re-use of data is in the multiple mini-batch input values for a layer (loop level 2) and the weights for a neuron (loop level 4). The same weight is used with multiple input values, and the same input value is used with multiple weights. This is akin to the reuse pattern in matrix-matrix multiplication, and the operation can be modelled as such and matrix-matrix multiplication code optimisation techniques can be used. This is shown in Figure \ref{fig:feedforward}.

The re-use for the weights in the simple algorithm description is carried by loop level 2, and the distance is the number of neurons multiplied by the number of weights per neuron. The re-use for the intermediate output/input values $a_i$ is carried by loop level 3, with the distance being the number of neurons in the layer. There is also reuse of data in the accumulation of the scalar $z$ value at level 4; this is in the inner loop and thus has re-use distance 1. Note that this is not a strict read-write dependency, but a tree-form accumulation for a reduction sum.

In general, the dependencies in the forward pass are from the output of a layer $L_i$ to the
input of the next layer $L_{i+1}$, meaning that the outer loop level 1 is essentially constrained to be sequential. The dependency structure in a given network can be more complex than this. Some outputs of neurons can be fed to later layers instead of just the next one, but these dependencies are less constraining than the immediate inputs to the next layer. Outputs of a layer also don't have to be fully connected to a following layer; it is possible for a sequence of layers to be composed of what are essentially different paths of separate unconnected sub-layers which share no dependencies. Thus there are no dependencies that impose as ordering across the different parts of different paths, thus giving some extra scheduling freedom in comparison to fully interconnected layers. This basic idea can be extended to quite complex DAG structures.


\paragraph{Backward Propagation:} The backward error propagation step is shown in algorithm \ref{algo:backward-prop}. Backward error propagation has a very similar structure to forward propagation. It can be viewed as the complement of it, with a slightly more complex scalar function being used which involves reading $z$ for that layer rather than just applying a function to a calculated value.

\begin{algorithm}
  \SetAlgoLined
  \KwData{$B$, a mini-batch of training points\\
    $M$, a model\\
    $z$, the weighted sums for each neuron (for each mini-batch input)\\
  }
  \For{all layers in reverse order (1)} {
    \For{each set of intermediate values from the mini-batch (2)} {
      \For{all neurons in the layer $L_i$ (2)} {
        \For{each weights $w$ from this neuron to a neuron in layer $L_{i-1}$ (3)} {
          Multiply the error $e$ of the neuron by the $w$\\
          Add the result to the running total $dcda$ for the that neuron in layer $L_{i-1}$\\
        }
      }
      \For{all neurons in the layer $L_{i-1}$} {
        Apply the differential of the activation function to the stored $z$ value for this neuron to get $v$\\
        Multiply the final running total $dcda$ by $v$\\
        Record the error $e$ for this neuron
      }
    }
  }
  \caption{Algorithm template for neural network backwards error propagation} 
  \hfill
  \label{algo:backward-prop}
\end{algorithm}

Consequently the dependency structures and reuse distances within the backwards propagation pass are the complement of those in forward propagation. Loop level 1 is sequential in the same way, the weights and intermediate values are accessed in the same way (matrix-matrix multiplication pattern) etc. etc.

\section{Insights and proof of concepts for increasing data reuse}
\label{sec:proofofconcept}

The experiments presented in this section have been conducted on a node from a 20 nodes cluster  equipped with dual 6-core Intel(R) Westmere CPUs with 12 hardware threads each, a clock speed 2.80GHz and 96 GB of RAM. The implementations are all sequential (executed on one core) and
C++ is used as a programming language. The results from the experiments are very much initial results given as an indication as to how some of the reuse and redundancy outlined in this document could be used.

\subsection{Cache-efficient SGD}

Caching is widely used in many computer systems because it increasingly impacts system speed, cost, and energy usage. The effect of caching depends on program locality, i.e. the pattern of data reuse. As a contribution towards increasing locality in current SGD and MB-GD algorithms described in Section \ref{sec:SGD} and hence caching possibilities, we introduce the Sliding Window SGD (SW-SGD). The objective of SW-SGD is to increase the locality of training point accesses while combining the advantages of SGD and MB-GD in terms of epoch efficiency and smoother convergence respectively.

\begin{figure}
     \centering
 \includegraphics[scale=0.5]{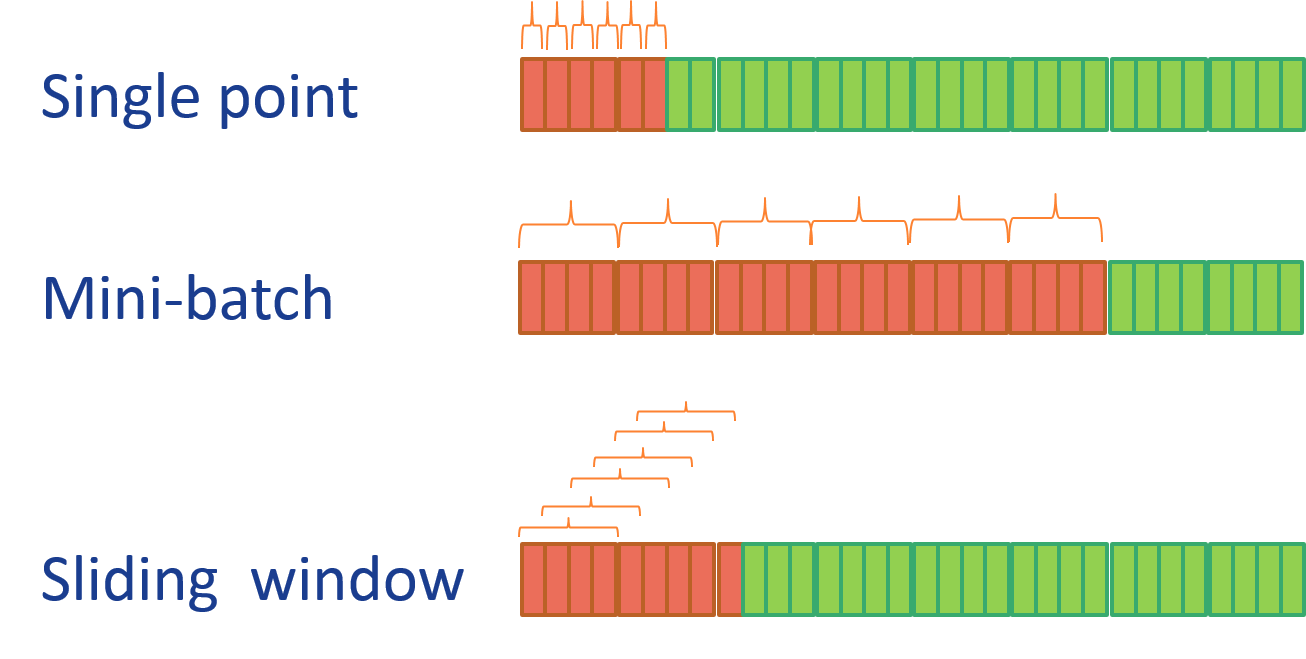}
 \caption{Comparing the data touched with six iterations of one point SGD, mini batch GD (MB-GD) and sliding window SGD (SW-SGD).}
 \label{fig:SWG} 
\end{figure}

As sketched in Figure \ref{fig:SWG}, the basic idea of the SW-SGD is to also consider recently visited points in the computation of the gradient. The list of recently visited points is kept in a vector potentially saved in the cache memory. The technique relies on the fact that computing the differentiated loss function on larger sized batches that come from cache is almost a free operation compared to loading new training points from the main memory. Indeed, as an example of how effective caches can theoretically be, let us assume a simple computation where access to main memory takes 40 cycles and access to the cache memory take 4 cycles (such as on Intel(R) Westmere CPUs \cite{Intel}). If the model uses 100 data elements 100 times each, the program spends 400,000 cycles on memory operations if there is no cache and only 40,000 cycles if all data can be cached. 

From machine learning and optimization prospective, using SW-SGD is also relevant since considering bigger sized batches in the computation of the gradient is likely to lead to a smooth and detectable convergence. First experiments towards this have been conducted on a classification problem of the MNIST dataset \cite{MNIST} containing 60,000 training and 10,000 testing images. The SW-SGD have also been tested on other gradient descent optimization algorithms such as Momentum, Adam, Adagrad, etc. (See \cite{SGDvariant} for a list of these variants and an entry to the literature.) The model to train is a neural network with 3 layers and 100 hidden units each. All the results are averaged from 5-fold cross-validation runs. A preliminary set of experiments was conducted in order to determine the best hyper-parameters (learning rate, batch size) of the algorithm. These parameters are used for the rest of the paper. 

\begin{figure*}
\center
\includegraphics[scale=0.4]{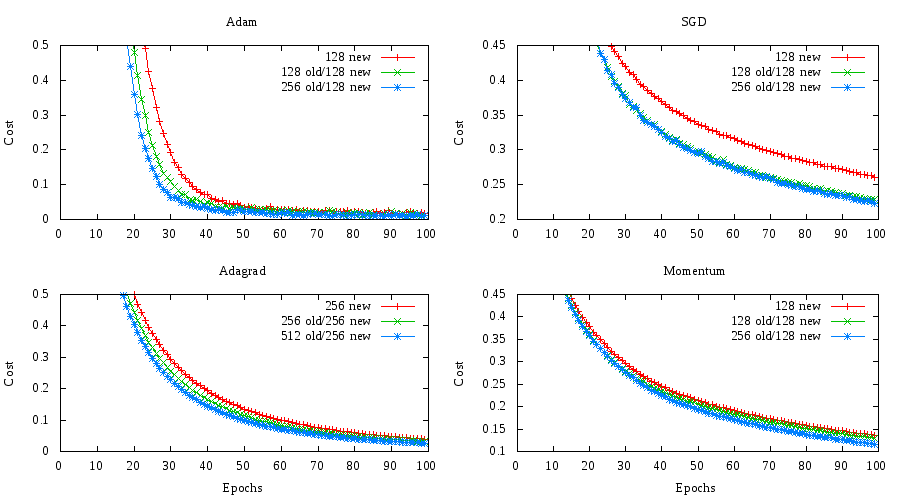}
\caption{Comparing different sizes of SW-SGD for different optimizers.}
\label{SWG2} 
\end{figure*}

In Figure \ref{SWG2}, different sizes of SW-SGD are compared for different optimizers. The aim here is to first prove that SW-SGD helps accelerating the convergence and second that it is orthogonal to the other gradient algorithms. We experimented with three scenarios for every algorithm: (1) only a batch of $B$ new points ($B$ being the best batch size from the preliminary experiments), (2) $B$ new points + $B$ points from the previous iteration and (3) $B$ new points + 2 $\times$ $B$ points from the previous iteration. 

For all algorithms, adding cached data points to the computation of the gradient improves the convergence rate. For example, for the Adam gradient algorithm, a cost of 0.077 is reached after 30 epochs when using training batches of 384 points (128 new and 256 cached) while at the same epoch (30) the first scenario where no cached points are considered the cost is 0.21 . Similar behavior can be observed with the other algorithms. This result proves that it should be possible to apply the fundamental idea of the SW-SGD to many GD algorithmic variants without any change to the definition of the algorithm. It is important here to highlight that, for the Adam algorithm for example, using a batch of size 256 and 512 new points is less efficient than using 128 new points as the first set of preliminary experiments showed that 128 is the best batch size. The added value here is therefore brought by the characteristic of considering old visited points in the computation and not because of a bigger batch size. 

One other possible direction we aim to explore in this context is to load different batches of points and to compute their gradients at the same time. Hence:

\begin{itemize}
\item future vectorisation opportunities are available.
\item the reuse distance is reduced for each point and as a consequence the temporal and spatial locality are increased.
\item matrix product best practices such as a cache blocking could be used where possible.
\end{itemize} 

\subsection{Coupling learners with similar data access pattern}

As mentioned earlier in this report, various machine learning algorithms exhibit very similar data access pattern and share a large volume of the computations in different steps of the ML process: training, optimization, sampling, etc. The two algorithms we discuss here are Parzen-Rozenblatt window and $k$-NN previously introduced in Section \ref{sec:IBL}. 

From a machine learning perspective, the Parzen window method can be regarded as a generalization of the $k$-nearest neighbour technique. Indeed, rather than choosing the $k$ nearest neighbours of a test point and labelling the test point with the weighted majority of its neighbours' votes, one considers all points in the voting scheme and assigns their weight by means of the kernel similarity function.

From a computation perspective, these algorithms similarly loop over all the points and sometimes calculate the same underlying distances (typically Euclidean). Therefore, the idea here is to run these two learners jointly on the same input data whilst producing different
models. A proof of concept has been made on a subset of the Chembl public data set \cite{CHEMBL} with 500K compounds and 2K targets. Our objective here was to give a first estimation of the amount of compute time that can be saved using the aforementioned optimization. In Table \ref{table:prw-knn}, the results are represented for two scenarios: both learners are run separately and the total elapsed time is accumulated and both learners run jointly. The elapsed times are in seconds. Two steps are considered: the time for loading the training and
testing sets and the time for the testing step (recall here that in instance-based learners no actual training phase occurs). 

\begin{table}[h!]
\center
\begin{tabular}{|c|c|c|}
 \hline
  & Load time (s) & Test time (s) \\
 \hline
PRW+$k$-NN separately & 7.545 & 2695.45 \\
 \hline
PRW+$k$-NN jointly & 3.726 & 1601.035 \\
 \hline
\end{tabular}
\caption{Comparing the elapsed time when running PRW and $k$-NN separately and jointly.}
\label{table:prw-knn}
\end{table}

The preliminary experimental results clearly show the added value of increasing data reuse when computing PRW and $k$-NN on the same pass over the data. The computing time is indeed almost divided by two.

\section{Conclusion}

In this research paper we have highlighted various access patterns with reuse and computations containing redundancy for a set of ML algorithms that we think are most used and useful. The analysis was conducted on these algorithms first in general covering sub-sampling techniques, multiple classifier selection and optimisation techniques. The second part of the contribution covered each class of the chosen algorithms specifically. We investigated for each the potential for performance improvements by identifying the reuse distances for the data used and the computations that could be grouped mainly between different learners.
This work is intended as a starting point for further work on improving the performance of learners as implemented on supercomputing clusters and applied to big data volumes.

\section{Acknowledgement}

This project has received funding from the European Union's Horizon 2020 research and innovation programme under grant agreement No 801051.

\end{document}